\documentclass[11pt]{article}
\usepackage{UF_FRED_paper_style}

\usepackage{lipsum}

\title{Applications of Machine Learning in Healthcare and Internet of Things (IOT): A Comprehensive Review}
\author{Farid Ghareh Mohammadi$^1$\\% Name author
    \href{farid.ghm@uga.edu}{\texttt{farid.ghm@uga.edu}} %% Email author 1 
\and Farzan Shenavarmasouleh$^1$\\% Name author
    \href{fs04199@uga.edu}{\texttt{fs04199@uga.edu}} %% Email author 2
\and Hamid R. Arabnia\\% Name author
    \href{hra@uga.edu}{\texttt{hra@uga.edu}}%% Email author 3
    }
    
\begin{document}
{\setstretch{.8}
\maketitle

\begin{abstract}
In recent years, smart healthcare IoT devices have become ubiquitous, but they work in isolated networks due to their policy. Having these devices connected in a network enables us to perform medical distributed data analysis. However, the presence of diverse IoT devices in terms of technology, structure, and network policy, makes it a challenging issue while applying traditional centralized learning algorithms on decentralized data collected from the IoT devices. In this study, we present an extensive review of the state-of-the-art machine learning applications particularly in healthcare, challenging issues in IoT, and corresponding promising solutions. Finally, we highlight some open-ended issues of IoT in healthcare that leaves further research studies and investigation for scientists.

\textit{\textbf{Keywords: }%
IoT, IoT Pipeline, Centralized and decentralized Big Data analysis, Online learning, Federated Learning.}
\footnotetext[1]{Authors contributed equally to this work.}
\noindent

\end{abstract}
}

% cite a few papers from each of them or maybe like a survey paper for supervised vs unsupervised
% talk a little about Spark and Hadoop that they can do most of it more efficiently. → not cnn though
% proofread and revisit titles

\section{Introduction}

\subsection{Internet Of Things} \label{iot}
Imagine being outside in a cafe and suddenly remembering that you forgot to turn off the stove before leaving the house. Traditionally, the only way to deal with this was to halt everything that you were doing at the moment and go back home to prevent hazardous scenarios. But, wouldn't it be easier if you could  address the problem remotely?
Well, you are in luck! Because in today's world Internet Of Things (IoT) has become a thing. IoT refers to a smart system of inter-related devices (computers, sensors, actuators, etc.) connected to the internet each with a unique identifier that are able to continuously communicate with each other with a common language over the network and collectively make intelligent decisions by analyzing the gathered raw data \citep{ashton2009internet, xia2012internet}.
% , extracting useful information and features, 

Nowadays, IoT is being used systematically to make human lives easier. This includes but is not limited to smart cities \citep{shenavarmasouleh2021embodied, amini2020promises, amini2019sustainable, mohammadi2021data}, smart farming \citep{jayaraman2016internet, pivoto2018scientific}, smart homes and grids \citep{komninos2014survey, jiang2018smart}, smart retail \citep{quintana2016improving}, and smart healthcare \citep{baker2017internet, li2021comprehensive}. Altogether, the automatic and ubiquitous nature of IoT increases the Quality of Service (QoS) and makes the entire system more efficient, more environmentally friendly, more sustainable.

\subsection{IoT in Healthcare}
Traditional monitoring in healthcare uses the resources and time insufficiently. A certified clinician needs to constantly check on the patient in-person and test results sometimes need days to get ready. Also, after hospital discharge, recovering patients may need to make a couple of appointments for the following check ups in order to make sure everything is on the right track with their health. 

With the advent of IoT, these issues are being addressed in a systematic way. Wearables and implantable devices are being utilized to continuously monitor the health of the patients independent from the time of the day and their physical locations. In addition to this, these devices can make use of local and/or more powerful centralized artificial intelligence (AI) models to not only detect but also predict diseases and hazardous scenarios. They can then automatically alert the patient and the corresponding doctor. We talk about the system pipeline in more detail in section \ref{iot_pipeline}.
Ubiquitous health (uHealth), electronic health (eHealth), and mobile health (mHealth) are all different names that cover the researches in this area and their ultimate goal is to reduce the cost of healthcare, increase patients satisfaction, decrease the load of the hospitals especially in the event of a crisis, and provide easy to understand yet accurate and powerful AI models to aid doctors in detecting and preventing diseases and providing personalized treatments
 \citep{shenavarmasouleh2021drdr, shenavarmasouleh2020drdr, shenavarmasouleh2021drdrv3}.

\subsection{Big Data in IoT Health}
Nowadays, millions if not billions of sensors are connected to patients that monitor and collect environmental, physical, physiological. and behavioral parameters non-stop. Recent trends also show the emergence of medical super sensors with more memory and processing power that can utilize Improved Particle Swarm Optimization algorithm to help accurate drug delivery to different organs of the human body to detect whether the drug has reached a particular position or not and much more \citep{sagar2020energy}. As expected these sensors generate a massive amount of data each second. This huge amount of heterogeneous data that contains highly redundant and correlated info is called Big Data \citep{sagiroglu2013big}. In the most naive way, all this data needs to be sent to a centralized server for feature extraction and analysis and this induces challenges such as network bottlenecks for transmitting the data and insufficient processing power and resources for real-time analysis of such data. Several solutions have been provided to solve the former problem such as removing redundant data and outliers in the local machine, aggregating the data before transmitting it, and trying to do a very basic analysis using light mobile AI models and only transmit data when the results hint to a problem \citep{ullah2021secure}. Machine Learning and Deep Learning techniques are the most dominant method for processing, understanding, and extracting knowledge from the collected data and improving the decision-making process, since after their training phase is done, they do not require any further supervision and can automatically perform their task.

% In this paper we ...

\subsection{Machine Learning for Big Data in IoT Health} \label{ml_models}
As mentioned above in section \ref{iot}, machine learning and deep learning techniques have been applied and are being used in all sorts of smart systems. These smart systems are comprised of many smaller components, each providing a very different service. Hence, if we want to understand smart systems we need to break them down into their building blocks. Computer Networks, Computer Vision, Natural Language Processing (NLP), Reinforcement Learning, and General Reasoning are among the main components. Each of these has its own separate world and an immense amount of ongoing research has been done over the past few decades. Hence, we would not go deep into them and try to touch on the most important architectures and algorithms that have been common in recent case studies.

In one broad view categorization, we can divide the machine learning algorithms into supervised and unsupervised techniques. In unsupervised techniques, the model receives unlabeled data, and hence its goal is to self-discover any sort of meaning and hidden patterns within the data to perform the data grouping. In more technical terms, this process is called clustering, and K-means, Fuzzy C-Means, Expectation-Maximization Algorithm, and Hidden Markov Model are some of the most prominent techniques. With supervised models, we can go one level deeper. They can further be split into classification and regression. 
In contrast to clustering, in classification, the model is given the labels in the training phase and it has to learn to categorize the input data in a number of pre-known classes with the least amount of error. K-nearest neighbor (KNN), Random Forest, C4.5 model, Naive Bayes, Support Vector Machine (SVM), Neural Networks, and Deep Belief Networks are a few of the most commonly used classifiers. When it comes to the images and videos, these classifiers cannot extract the best features all on their own and will be in need of help. Convolutional Neural Networks (CNN) address this issue using their pooling and convolutional layers and have been being used for any task performed on media since AlexNet \citep{krizhevsky2012imagenet} won the ImageNet image classification competition in 2012. They have been a tremendous amount of research on CNNs and the models that are being used nowadays are several folds stronger than the initial model. This is all thanks to researchers who thoroughly studied and analyzed CNNS which their efforts ultimately led to expanding and improving these networks. VGG \citep{simonyan2014very}, Inception \citep{szegedy2015going}, and ResNet \citep{he2016deep} are some of the best CNNs that are being used recently to handle all kinds of vision problems, namely, Image classification, Object Localization, Object Detection, Semantic Segmentation, and Instance Segmentation. Also, if temporal features are of importance, then Recurrent Neural Networks (RNN) such as Long short-term memory (LSTM) \citep{hochreiter1997long} or Gated recurrent unit (GRU) \citep{cho2014learning} can be employed on top of the CNNs to learn the time-related patterns. CNN architectures mostly include a simple classifier such as KNN or a feed-forward neural network as their last few layers to perform the classification after successfully extracting features using their convolutional layers, but after the training, the default classifier can be detached and the convolutional layers can be solely used as feature extractors and they can be fused with other features and be fed to other classifiers or ensembles of them as needed \citep{asali2021deepmsrf}.

While classification tries to assign a label to a given input, regression aims to predict a quantity in the continuous space. Hence, these models are useful for predicting numerical values. Linear regression, Support vector regression (SVR), and a properly structured neural network are among the to-go algorithms for this task.

\begin{figure}[t]
    \centering
    \includegraphics[height=1.7in]{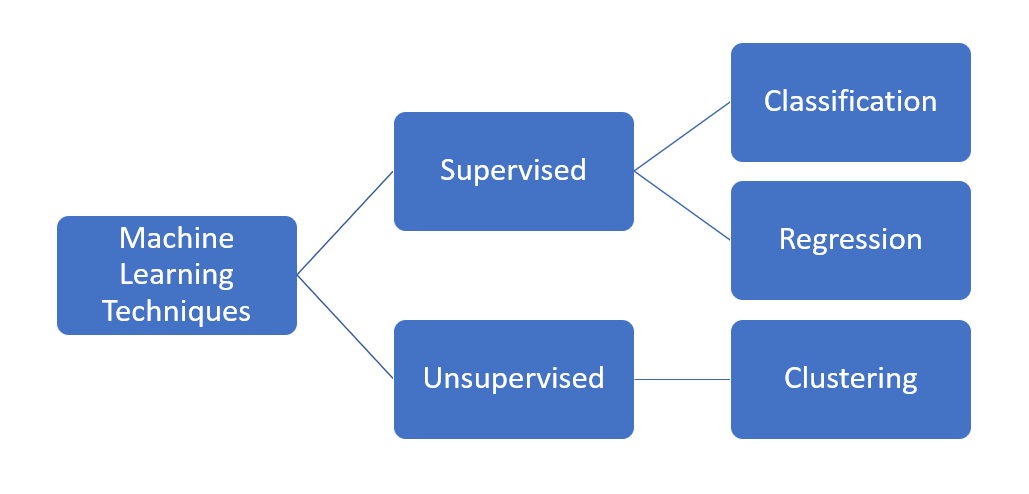}
      \caption{Overview of Machine Learning Techniques}
        \label{fig:mlfig}
\end{figure}
 
\section{IoT Pipeline} \label{iot_pipeline}
\subsection{Sensing and Data Gathering}
IoT aims to automate tasks that otherwise need to be tackled manually and by doing so improves the quality of lives of humans, and in the context of healthcare, this would mean ubiquitous monitoring and treatments of patients with improved responsiveness and accuracy. 
Advancements of portable microcontrollers and microprocessors such as Arduino and Raspberry pi, and the emergence of efficient, low-cost, yet accurate sensors and communication modules such as Zigbee that could easily be integrated with these devices led to the rise of wearables and implants. Wearables, as their name suggests, are products that could be worn as accessories. They come in various types, sizes, and shapes, but smartwatches, fitness trackers, bracelets, and smart rings are considered the most prominent ones. They can track and monitor heart rate, blood pressure, workout time, calories burned, sleep time, and much more. Implants, on the other hand, are devices that have to be inserted into the body or placed under the skin of people. Implants are used for applications such as heart pacemakers \citep{difrancesco1993pacemaker}, hearing restoration \citep{schmerber2017safety}, glucose monitoring \citep{csoeregi1994design}, and spiral neurons \citep{leake1999chronic}. Aside from monitoring vitals, implants usually have a more dedicated goal and aim to help a particular organ in the body to restore its biological system. For instance, in the case of glucose monitoring, the sensor measures the amount of glucose in the blood regularly and also has the ability to regularize it by injecting insulin from its embedded insulin tank as needed.

\subsection{Data Storage and Preprocessing}
Real-time transmission of gathered data is often costly and hence usually data collected by wearables and implants will be stored on a local memory on the device for initial analysis and preprocessing. The aforementioned models in \ref{ml_models} often require a lot of processing power and working memory and are not able to perform well on resources embedded in wearables and implants. So, lighter neural network models are employed to address this. They often sacrifice a bit of accuracy for faster and less resource-hungry computations and are getting better day by day. Trained versions of pruned or reduced neural networks \citep{blalock2020state} such as SqueezeNet \citep{iandola2016squeezenet}, MobileNet \citep{sandler2018mobilenetv2}, and EfficientNet \citep{tan2019efficientnet} can be employed in such devices to perform general analysis on data streams and even video streams collected from small cameras on some implants.

\begin{figure}[t]
    \centering
    \includegraphics[height=1.4in]{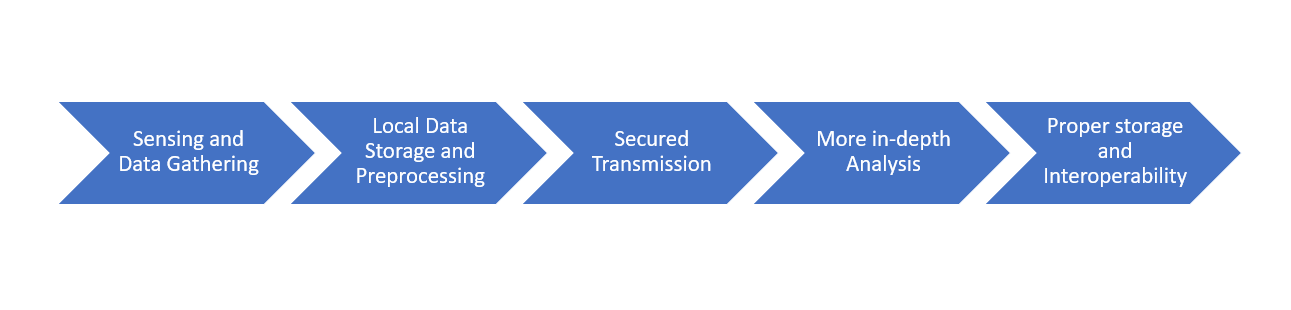}
      \caption{Overview of IoT pipeline}
\end{figure}

\subsection{Secured Transmission}
With the presence of preprocessing models in devices, the transmission of data to centralized servers can be done in an asynchronous and passive way. These models analyze and keep track of patterns in streamed data and if no anomaly is detected, then the data will be saved and transferred at a later time or on a fixed schedule when the network is less busy. But, if the models detect, suspect, or predict a problem with the underlying data, they can proceed and immediately transfer the required information for further analysis. 

Even though preprocessing models help a lot, but still they have limited computational power and cannot do heavy analysis using the embedded battery and processor and without access to the data from other patients. Hence, the entire data needs to be transferred to the server at some point and that proposes two main problems. First, different sensors generate a lot of heavily correlated data, and altogether the size of generated data is big. So, it would be a good idea to reduce the size of data as much as possible before actually transferring them. Here, again different light local models can be incorporated to detect and remove outliers and noise and remove redundant data. The remaining data can be then get aggregated and compressed and broken down into chunks to be sent to the central server when the load is low \citep{hussain2020machine, wan2019similarity}. There is also a lot that can be learned from swarm intelligence and how the data transmission is being addressed in that field \citep{wang2018bandwidth}.

Second, the transmitted data includes a lot of information about the patient, and secured transfer must be taken into account to prevent data leaks, eavesdroppers, DoS attacks, and much more caused by misconfigured network and communication devices. It goes without saying, that the data itself should be encrypted to maintain the confidentiality of the patients even in the event of network breaches. Hence, security standards and techniques are constantly being enforced to avoid these scenarios at all cost \citep{gope2015bsn, jan2019payload, khan2018performance}.

\subsection{More in-depth Analysis}
The main strength of central computing servers is their access to a relatively large amount of resources and their ability to make use of the computing power to deeply analyze and find patterns in the data. Here, the machine learning algorithms access to data goes above and beyond just one patient and they could leverage all the information from different patients at the same time by fusing them and trying to find patterns and relations on a much higher semantic level. Given their more horsepower, central server analytical frameworks can also merge any possible historical information that the patients may have from their clinical visits and/or by using different services and sensors from other providers, their demographic information, and other more complementary data before performing the analytics to get the complete picture more easily.

Deep learning and machine learning techniques are perfect for analyzing the data at hand, but sometimes we could only perceive them as black boxes, and even though this might not be an issue in many fields but it is not acceptable in healthcare. The results and the corresponding decisions that would be made using those models must be interpretable by doctors and clinicians since computers are here to aid the healthcare community and for making it possible to work side-by-side with them, they have to find a way to elaborate their complex decisions especially if they happen to be the incorrect ones.

Machine learning algorithms have an easier time achieving this since most of the time their logic is not inherently very complicated and is easier to follow. Decision trees for instance can easily illustrate their "decision tree" and show all the underlying conditions and attributes that lead them into making a particular decision. Even if the number of features is hundreds of dimensions, they are still ways to reduce the number of dimensions and show the results of methods such as KNN in relatively accurate 2-3D plots \citep{mohammadi2021evolutionary, cao2003comparison}.

However, given the complex architecture of deep learning models, this would not be a trivial task and the decisions cannot be easily traceable. To tackle this issue, there is ongoing research on how to generate rules from neural networks and trying to find the most dominant paths in them to make some sense out of their decision \citep{wang2018interpret, fu1994rule}. Also, when it comes to processing images and videos, attention models can be used to visualize the areas that the model mostly focuses on while deciding on a particular problem \citep{wang2017residual}.

 \subsection{Proper storage and Interoperability}
The newly received data should be properly stored for future use cases; however, it is not as easy as dumping the entire data into a database. As mentioned, central frameworks can lend or borrow information from other central frameworks and services. Yet, there is no guarantee that the other servers use the same standards for data formats and some norm such as interfaces or database level rules and schemas needs to be incorporated. Hence, semantic interoperability is a must while dealing with multiple computer systems exchanging and using each other's information which if we look at it again is the core concept that the entire IoT is built on. Every data must have meta-information about different entities within it to give context to the corresponding values and bring in the possibility to connect and link these little pieces of data and do some automatic reasoning and inferring to ultimately transform data into knowledge. Resource Description Framework (RDF) \citep{miller1998introduction} and Web Ontology Language (OWL) \citep{mcguinness2004owl, antoniou2004web} are two very different but complementary ways to perform such tasks and query languages such as SPARQL \citep{perez2009semantics, barbieri2009c} can be used to query such data from different sources all around the internet. Semantic interoperability is a very big field with numerous standards and techniques that are beyond the scope of this paper so we would spare the details for the sake of this review.

% ----------
\section{Some of the Applications of IoT in Healthcare}
Machine learning and deep learning is being used in many different parts of health care such as the discovery of new drugs, manufacturing them, aiding doctors in surgeries, performing radiation treatments, and much more. But, when it comes to IoT, the primary use cases of them would be in developing remote monitoring, prediction and recommendation systems, and living assistants. The borders between these systems are not as clear-cut as you might imagine, since almost all the papers that will be discussed in this section can fall under two or more of the categories. Hence, to avoid redundancy, we abstain from categorizing them.

First and the foremost is ubiquitous monitoring using sensors embedded in wearables and implants because without them there would be no data to analyze. \citep{shaikh2012real, corchado2009using, kim2015coexistence, ingole2015implementation, daugtacs2008real, guo2008long, vippalapalli2016internet} all propose different telemonitoring systems with varying configurations but with the same purpose. Their systems aim to monitor and investigate the health of patients whether ordinary, critically ill, elderly, or recently discharged from hospital by tracking parameters such as heart rate, blood pressure, body temperature, ECG, EEG, EKG, and other physiological attributes and sending them to central servers using various wireless technologies like wifi, Zigbee-based WBAN, WSN, and BSN, altogether enabling the doctors to monitor the health of the patients from the convenience of their rooms.

In a more unique use case, authors in \citep{gondalia2018iot} propose a system to track the location and monitor the health and condition of war soldiers on the battlefield that in return makes the search and rescue operation faster and more accurate in the event that an individual gets injured. Aside from the aforementioned physiological parameters, they also use a humidity sensor, vibration sensor, bomb detector and then use a hybrid of ZigBee and LoRaWAN network infrastructures to transfer the data to the control room either after some fixed interval or only when there is a significant change in conditions of a soldier. In \citep{nikam2013gps, pramod2014gps} authors also present ways to track the health of soldiers as well as the ammunition on them.

All the collected data would be then get merged with possible demographic information and historic sensory data and get thoroughly analyzed using different AI techniques to predict if everything is on the right track and the patient is healthy or whether something is wrong and require immediate attention. For instance in the case of soldier data, in \citep{gondalia2018iot}, the authors apply K-Means classification to the data and classify individuals into healthy, ill, abnormal, and dead. Authors in \citep{khan2019mobile} use the data to detect whether individuals are stressed or not. \citep{yao2019secured} and \citep{kumar2018novel} use ECG and other sensory data for earlier prediction of heart-related diseases. In \citep{ganapathy2014intelligent} and \citep{yadav2019machine} authors utilize different fuzzy rules and techniques such as Probabilistic Fuzzy Random Forest (FRF) to classify and efficiently predict various diseases. The latter also shows that FRF models perform better than Linear Regression and Q- Learning algorithms in their study. Hybrid and more general frameworks have been also proposed in \citep{kirtana2017iot} and \citep{verma2019comprehensive} to detect and monitor several diseases such as diabetes and cardiovascular-related ones and forecast their severity at the same time.
In \citep{otoom2020iot} authors proposed to use IoT frameworks to identify transmissible diseases at their early stages and prevent them from turning into a crisis.

% %%%%%%
% In \citep{ali2018type}, the authors proposed a novel recommendation system based on Type-2 fuzzy ontology-aided RS,
% %%%%%

Living Assistants are without a doubt one of the main use cases of IoT and are assisting millions of patients, especially the elderly population on a daily basis by providing personalized analysis for disease prevention and collaborative care, all thanks to the advancements of wearables, implants, AI models, and communication networks. We have covered most of the services and functionalities that wearables and implants can provide in earlier sections, but aside from them, authors in 
\citep{li2020secured, kinthada2016emedicare, alex2016modern}, propose frameworks and tools such as home-based wireless medical boxes to track and monitor medicine consumption of patients, help them in categorizing of drugs, keep logs of intake history, and warn them if the pill is not taken on time. Of course, this can also be synced with vital sensors of the patient wearables and if a significant change occurs due to drug consumption, the corresponding doctor can be notified immediately. It goes without saying that this would be a huge help for our elderly citizens, especially the analphabets.

Also, most medications must be maintained at a constant condition and temperature. \citep{basingab2020distributed} suggests that millions of dollars get wasted annually due to random and unexpected refrigerator failures. So, the authors in \citep{basingab2020distributed, mambou2016monitoring} propose IoT techniques to constantly monitor the condition of drugs and prevent costly errors.

\section{Machine learning Challenges in IoT  Healthcare}
\begin{figure}[H]
    \centering
    \includegraphics[height=2.5in]{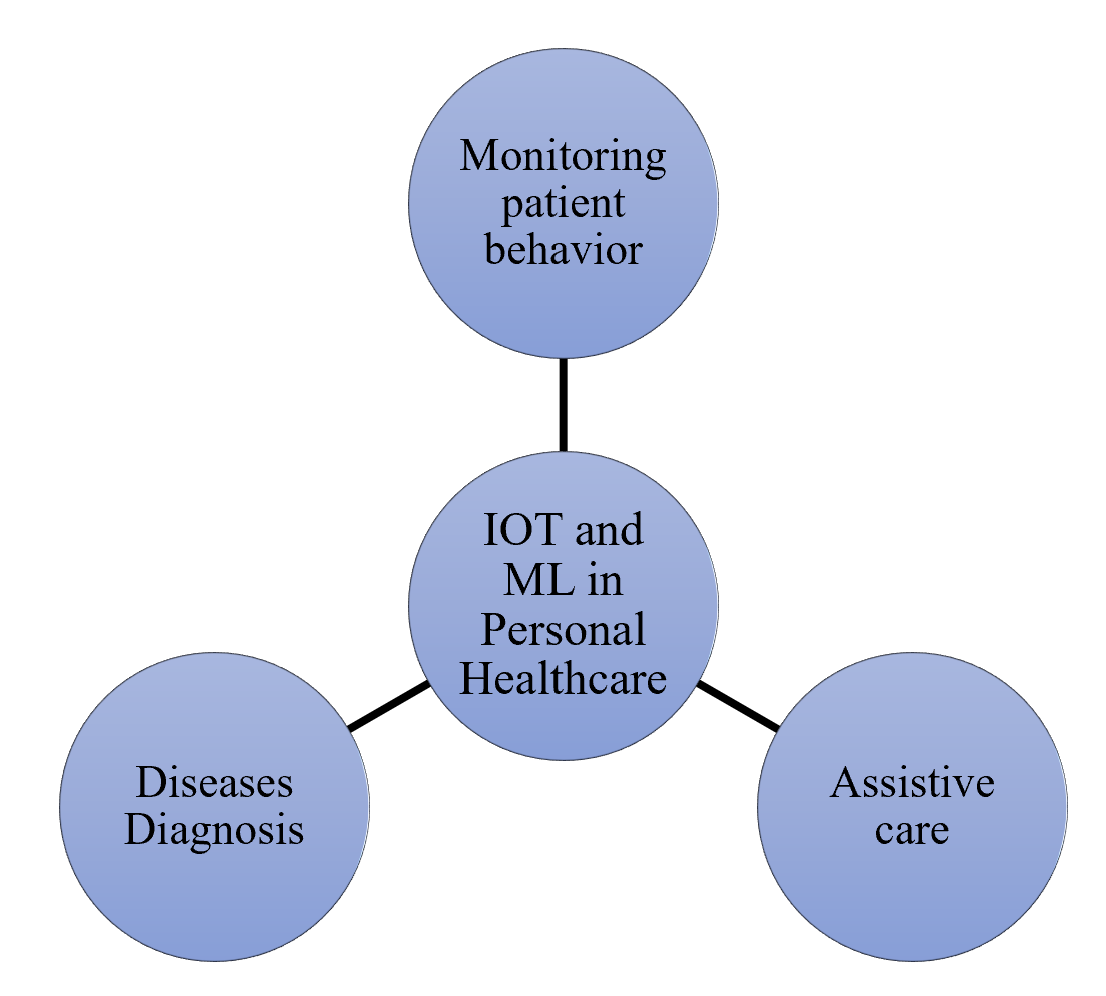}
      \caption{A general schema of IoT and Machine Learning applications in Personal Healthcare (PH) }

    \label{fig:IoTML}
\end{figure}

In this section, we mainly list and discuss the major limitations and challenges of machine learning in IoT healthcare. Figure \ref{fig:IoTML} presents a tight relationship among IoT, machine learning (ML) and personal healthcare(PH). There are many studies that state the applications of machine learning in IoT and PH \citep{ahamed2018applying}. The figure shows that IoT generates data that feeds ML algorithms and then the outputs provide solutions for PH such as disease diagnosis, patient behaviors analysis, and assistive care advice.

ML and IoT-based assistive PH services have already had and will have so many impacts on peoples' lives due to technological advancement. However, assistive PH will require to face challenging issues such as usability and affordability \citep{shahrestani2017assistive}. Additionally, privacy and authentication problems in IoT devices can attract hackers' attention and cause issues as they will be hacked if not correctly secured \citep{shahrestani2017assistive, ahamed2018applying}.

Further discovery shows that using ML-based PH service enables us to use a predictive analysis approach that assists released patients from a hospital who may need to readmit to the hospital. The goal of predictive analysis is to create a risk classification model in which particular patients with higher risk are controlled with additional effort and assistive care such as providing additional monitoring IoT devices/ sensors and constant (real-time) follow-up and analysis. These models are highly generated based on past historical experience and data. The dynamic PH system which would assist in re-admission avoidance initiatives must also leverage dynamic data from the patient and take it into consideration to predict future possibilities and initiate an action plan to mitigate probable complications \citep{ahamed2018applying}. 

\begin{figure}[H]
    \centering
    \includegraphics[height=2.5in]{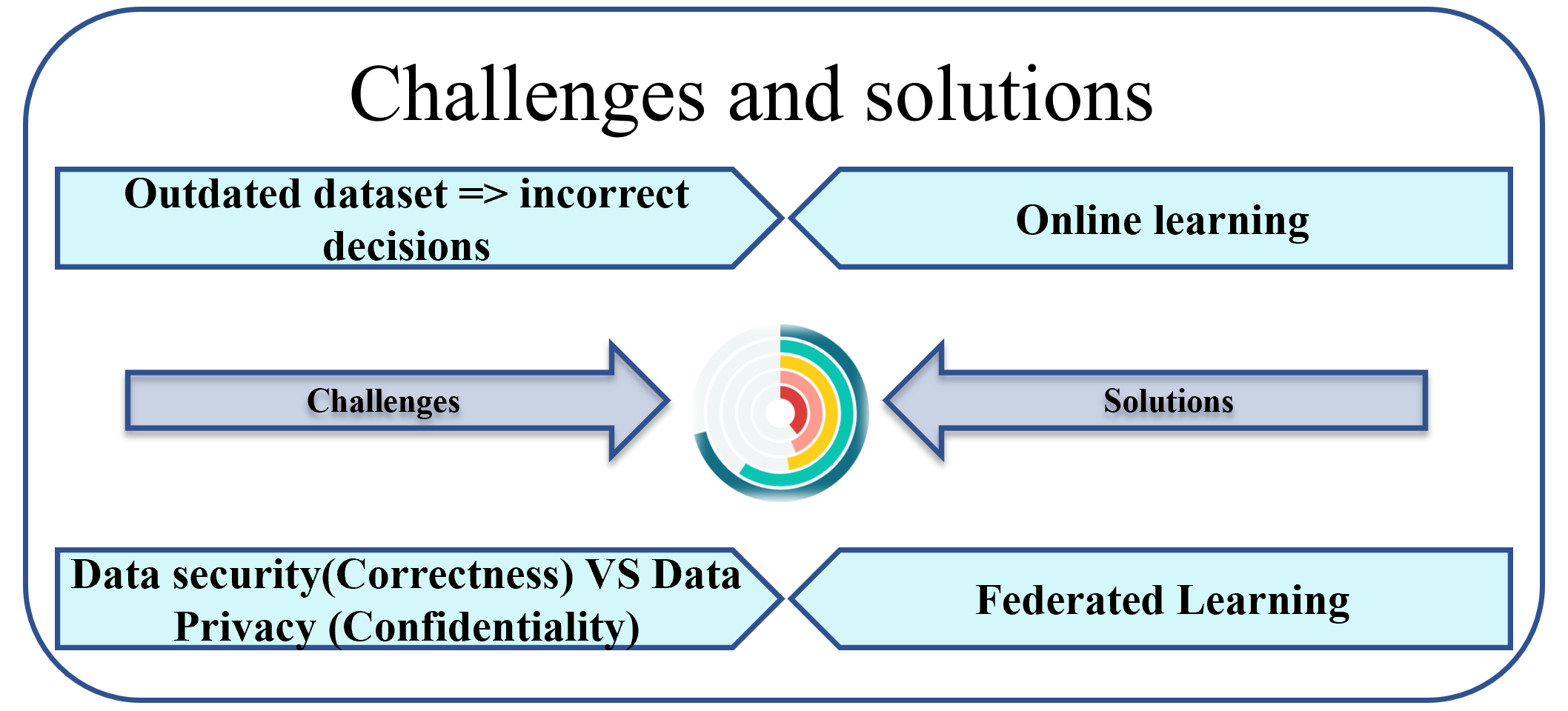}
      \caption{A general ML challenges in IoT healthcare and associated solutions}

    \label{fig:glch}
\end{figure}

We present an abstract view of challenges and solutions in figure \ref{fig:glch} which shows two challenges: 1) outdated dataset which leads us to incorrect decisions, and 2) Data security and data privacy which lower the reliability of IoT devices. On the other hand, the figure provides us two associated solutions: 1) Online learning to keep learning for the new entry data and 2) Federated learning for learning from distributed data among end-users. In the current section, we explain the first challenge and the second one by elaborating on them when they may occur. In the next section, we elaborate on the corresponding solutions for giving two solutions. However, the solutions also have their own pros and cons.

%\subsection{Security}
%Having a variety of devices from simple objects to smart objects, IoT is getting popular in the majority of industries because of Artificial Intelligence’s rapid progress. This has a long-term yet important influence on the health monitoring, managing, and clinical service to patient’s physiological information \citep{selvaraj2020challenges}. 
%Patients are highly connected with different objects of sensors and the data generated from the sensors will be forwarded to the health-monitoring unit. Sometimes data are recorded in the cloud, which helps to analyze the number of data with security. 

%\subsubsection{Challenge and Solution number 1}
%An important topic in every domain, particularly in the IoT is security due to data transmission. When facing data transmission from the sensor to a cloud center, there is highly likely a possibility of losing integrity and confidentiality of data, and additionally, it is not a trivial task to encrypt the data received from low-resource devices. \textbf{Solution:} A distributed environment namely Cloud is needed to store the medical data which provides more flexibility for remotely caring patients accessed by doctors and patients. The IoT and cloud start performing real-time processing which ends up having high complexity problems in architecture for sending and receiving data. 
\subsection{Challenge $\#$1 : Outdated dataset}
\label{section:Chl1}
There have been many research studies working on applications of machine learning (ML) in IoT healthcare. The ML algorithms develop analytic models integrated into diverse healthcare service applications and clinical smart systems \citep{selvaraj2020challenges}.
These models mainly get evaluated on the collected data from IoT devices to recognize behavioral patterns and different clinical conditions of the patients such as identifying the patient’s
improvements, habits and their anomaly actions in daily routine activities, different behavior of sleeping, digestive, drinking, and eating pattern.  Having those patterns available, the smart decision-making systems recommend particular lifestyle advice, care plans, and special treatment for the patients. Additionally, the doctors can further be engaged in the care plan process to evaluate and validate the lifestyle advice and care plans.  Medical data including clinical, lifestyle, behavior are so sensitive and it is highly likely that there may be different types of biased engaged in the process of data collection and may not be diverse enough to govern all scenarios. Furthermore, the noisy, incomplete data could lead to a lower probability rate to detect and predict a health-related diagnosis and advisory notice.

The training dataset and generated model could have an outdated version of the dataset and even if we had a rich model, it will not be effective anymore. Having old and outdated models and datasets, lead us to an incorrect decision derived from the smart systems \citep{ahamed2018applying}.

\subsubsection{ML challenges in Assistive Care }
%\citep{selvaraj2020challenges} % Example of citation. Erase before use
We believe that ML algorithms are highly related to statistical analogy and deduction in which the ML algorithms make decisions and predict using existing and previous experience (training dataset). In the case of monitoring a patient, the ML-based method will monitor and analyze the situation according to the training dataset. Thus, the training dataset plays an imperative role in seeking the current pattern and predicting the future trend of a given new problem during a test phase. This dataset sometimes is biased and would not be as diverse as possible to cover many scenarios. For example, let's consider a sleep monitoring case, sleep patterns can vary from person to person, from kid to elderly and health status. Therefore, a complete dataset of all case studies are not existing during training time to keep a track of sleep patterns, and this may lead us to an incorrect estimation in PH \citep{selvaraj2020challenges}.
 
Furthermore, using IoT and ML enable PH to make a decision for diagnosis, prediction. There are some examples, in which ML-based decisions could not be correct, and it is not possible to state why a certain decision was made. For instance, in the case of autonomous cars, a few accidents occurred due to wrong decisions made by the cars \citep{ahamed2018applying, selvaraj2020challenges}. 

The main point here is how to assess the decision taken by an AI machine when unsupervised machine learning algorithms were used. This may lead us to an ethical question of who would be responsible in case of a false statement and how to recognize or rectify that incompleteness in the process of decision-making and get to know how would unsupervised machine learning algorithms work. These challenges would limit the usage of ML algorithms in using PH for a sensitive application, particularly personalized medicine care \citep{selvaraj2020challenges}. 

\subsection{Challenge $\#$2: Confidentiality $\&$ Correctness }
\label{section:Chl2}
The invention of medical Internet of Things (IoT) devices and the high popularity of ubiquitous wearable devices enables us to have non-stop personal healthcare monitoring everywhere such as home, work, and hospital environments for different applications like childcare and assisted living. These devices record real-time electronic health measurements from a variety of objects and sensors and transfer these patient data to an application server to be pre-analyzed and pre-processed to restore on a data server \citep{jourdan2020privacy}. 

These processing and analysis use machine learning algorithms to provide different services such as motion tracking, presenting the number of steps, burned calories, sleep monitoring, traveled distance, and essential signs measurements like heart rate, electrocardiogram (ECG), skin temperature, and electroencephalogram (EEG) \citep{haghi2017wearable}. So, analyzing the data generated by IoT devices and sharing them through a connected network to a server raise uncertainty, trust, and confidentiality issues. The data stored on a server and the communication networks are vulnerable to breach. So, data security and privacy are important challenges to take into consideration. There are basic solutions provided to increase data security by applying encryption algorithms and keeping the data safe however, if a hacker gets to know the key to the decryption algorithm and uncovers the message, then the confidential information will be everywhere. Furthermore, while encryption, there is a possibility of losing some information and if the decryption algorithms fail to retrieve all original data, the process of encryption and decryption is not useful anymore.

\subsubsection{ML Challenges in Monitoring Patient Activities }
In a certain case of monitoring patient activities, a number of research studies take advantage of data generated from motion sensors to define physical activity in real-world settings for a variety of case studies. Research studies have proved that motion sensors type is one of the reliable tools to assess long-term human physical activity for particular cancer patients during the time they are in their therapy sessions \citep{gupta2018feasibility}.

However, due to IoT devices features, gathering data from medical IoT sensors/devices for healthcare application are very sensitive. Recent advances in web technologies and wireless communication/transformation enhance the gathering data process and the remote real-time monitoring \citep{jourdan2020privacy}. But, the complicated workflow of gathering medical data increases the security issues and privacy risks during the life-cycle of the data gathering such as the data collection and transmission, and additionally the processing and the storage \citep{rushanan2014sok}.

In the activity recognition process using mobile devices rather than wearable devices, the challenging issue relies on determining the data that can preserve the privacy of users while it is relevant and important for machine learning tasks \citep{jourdan2020privacy}. To address this challenging issue, we may face and answer two questions: first, does the gathered data has high protection layers or encryption protocol so that no one is granted access to it? Second, how to assess if the protected data is maintained as accurately as captured? Obtaining a trade-off between data computation and data privacy is an essential objective to have a secure transfer of data and protect data using mobile devices and enhance the end-users trust.

One of the main challenges of using IoT for healthcare monitoring is the security, particularly privacy of patient information in the machine learning analysis process. An efficient user authentication framework ensures that only legitimate account users have access to data and services. So, the problem is the data accessibility for users that is vulnerable for hackers to access sensitive.  Sharing information from IoT devices as it is will be problematic \citep{jourdan2020privacy}.

\section{Alternative Promises in IoT  Healthcare }
The IoT usage statistics have shown the potential application to many medical and healthcare domains with access to large volumes of corresponding data gathered by IoT sensors/devices. However, the increasing need for high healthcare data security and data privacy forces  IoT devices to be known as a disconnected island of data \citep{yuan2020federated}. Furthermore, due to lack of updated data model issues, data privacy, and security which are the main challenges using traditional machine learning algorithms mentioned in section \ref{section:Chl1} and \ref{section:Chl2}, we investigate advanced machine learning promises with IoT-healthcare applications in this section. We elaborate on the promises in the following sections. First, we discuss the online learning algorithm which addresses the lack of updated data and provides a solution in which the classifier can learn from upcoming new data and update the model in real-time on each learning iteration. Second, we present a federated learning technique that enables us to learn from distributed data collected from end-users without transferring the collected data to an application server.

\subsection{Online Machine Learning or Adaptive Learning}
A basic procedure of a machine learning algorithm is already discussed and shown in section \ref{iot_pipeline}. Additionally, we discuss in section \ref{section:Chl1} what are the challenges of traditional machine learning algorithms. Thus, in this section, we aim to propose a solution that enables scientists to address the challenge properly by using an online learning algorithm.  

In machine learning, there is a fundamental algorithm called online learning or adaptive learning in which we feed data in a sequential order which is highly recommended and used to update the best classifier for future data even if the training process is done. In comparison to batch learning algorithms which generate the best classifier by learning only from the training dataset and never get updated afterward. In other words, online or adaptive learning consider tasks (during the training process and after that) arriving in a stream rather than an offline finite dataset. But, the tasks are associated with the ability to efficiently adapt to the current task in the stream with respect to the corresponding learning rate, more than remembering the old tasks \citep{hassanreal}.

Furthermore, online learning is a new technique in machine learning where it is physically and logically infeasible to train over the training data such as patient data over therapy sessions, requiring the need for data will be generated in the future. It is also used in conditions when it is essential for the algorithm to train whole data due to lack of memory \citep{fontenla2013online}.

\subsubsection{General Process of Online Machine Learning}
\begin{figure}[H]
    \centering
    \includegraphics[height=2.75in]{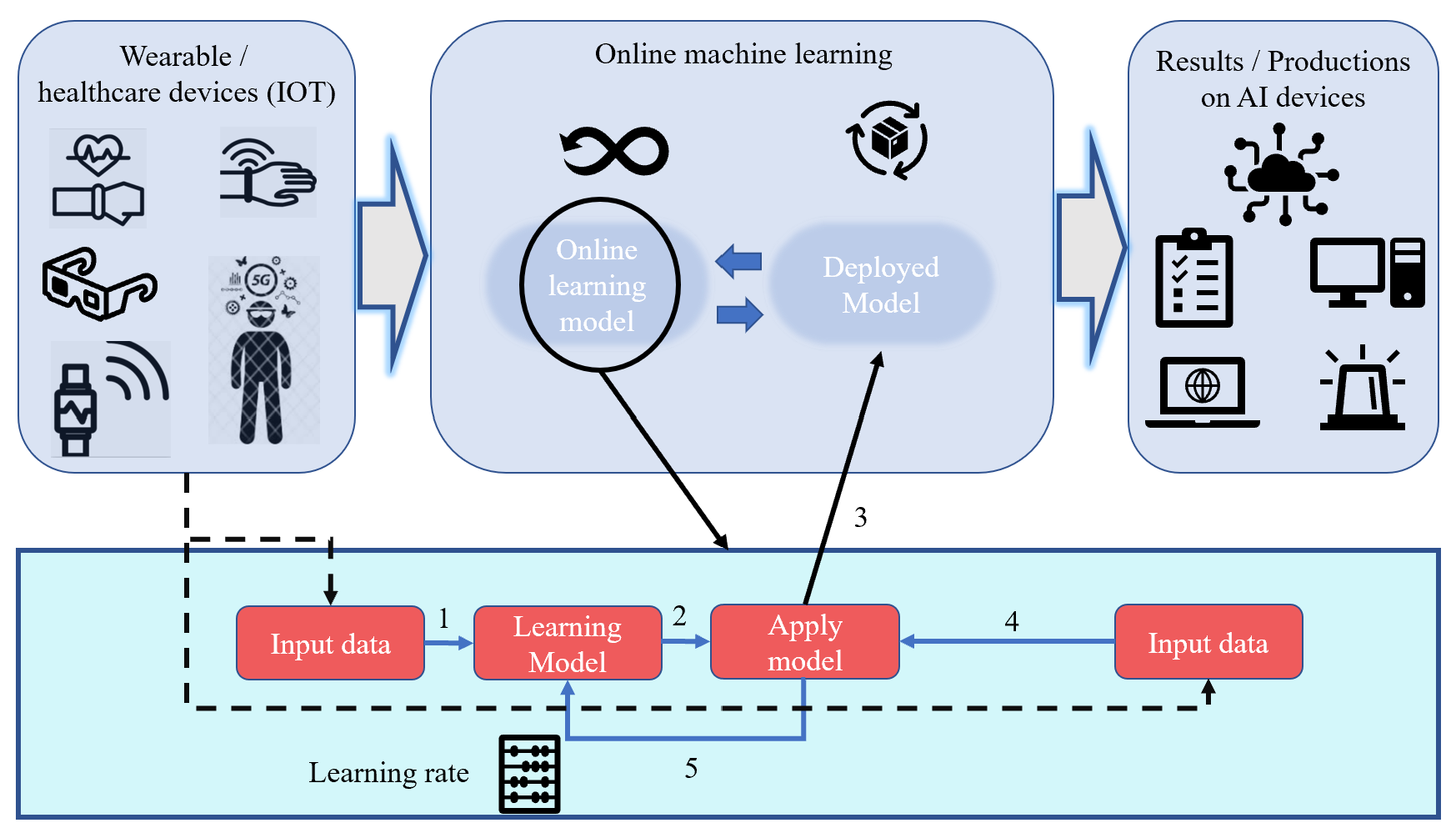}
      \caption{A general schema of Online Machine Learning }

    \label{fig:OnlineML}
\end{figure}

In this section, we aim to provide a very brief overview of ab online machine learning algorithm. Figure \ref{fig:OnlineML} presents a general schema of the online machine learning algorithm. In general, the online machine learning schema works with three main packages: 1) Input package including wearable or healthcare IoT devices, 2) Online machine learning package involves of 5 steps, 3) The results and predictions projected on AI devices. The main online machine learning steps are the same traditional machine learning process, except one more step which is re-training the learning model with a learning rate that denotes the importance of new input data. The larger the learning rate, the more important input data is considered. However, if a larger learning rate is chosen, this model will be highly likely exposed to outliers and may end up stuck in local minimum or maximum. On the other hand, the smaller the learning rate, the less exposed to outliers. However, it takes a long time to get the model updated based on new entry data due to the smaller learning rate.

The five steps in the online machine learning package presented in figure \ref{fig:OnlineML} play an important role in the process of applying online learning in IoT healthcare.  Within these five steps, steps 1 to 3 are the typical learning process in traditional machine learning including reading input data (training data, training a model, and applying the model). Online learning algorithm starts with step 4 where the algorithm accepts a new sequence of data and in step 5  the algorithm learns from the newly added data and updates the model with a learning rate customized value.

\subsubsection{Online Learning Applications in IoT Healthcare }
With advances in the world of technologies such as artificial intelligence, particularly in machine learning and IoT devices, the AI applications and IoT objects usage have proliferated in many domains, more specifically and demanding in medicine and healthcare. Adaptive learning, also known as never-ending learning \citep{mitchell2018never} or online machine learning \citep{fontenla2013online}, is a new basic idea in machine learning in which models learn from data continuously and evolve based on the new input of data with a certain learning rate while maintaining previously learned knowledge. This non-static process of supervised machine learning algorithms allows the model to iteratively learn from data and automatically update its behavior with the learning rate. The smaller the learning rate, the smaller range of behavioral we will have. In other words, we do not add more weight to the upcoming input data. However, we aim to have the model updated sooner based on the new data, we may need to initialize a pretty large value for the learning rate which will not get affected by outliers. One of the examples of this online learning are recommender systems utilized by companies such as Netflix and Amazon, however, such systems have their own challenging issues like fairness in the machine learning process \citep{ahmad2020fairness}.\\

\textbf{Big Data streaming:}

A big data streaming computation has been highly applied as a fundamental role in real-time healthcare analytics. There are applications of online machine learning such as real-time monitoring and tracking systems which play an essential role in the healthcare and medical domain. Mobile applications and sensors and wearable medical devices/sensors are the examples of popular rich sources that have been generating constantly a high volume of data namely, streaming data \citep{hassanreal}. Apparently and logically, using traditional machine learning algorithms to apply on the streaming data to make real-time actions in case of emergencies looks a difficult task. Therefore, it is highly required a solution for real-time big data processing to make sure the results are effective and explainable. For example, a real-time solution for monitoring flu and cancer patients is proposed by applying Twitter mining tool \citep{lee2013real}. Another big data model for real-time medical analysis is investigated in \citep{akhtar2016challenges}. The model used Spark streaming and Apache Kafka to get evaluated on a stream of healthcare data. In \citep{ed2018application} a real-time health status prediction solution was proposed by applying Apache Spark, which is a powerful big data analysis tool. Spark was used successfully for streaming data due to traditional machine learning limitations and handling distributed computations.

The second important extension of online machine learning is online meta-learning. Finn \emph{et. al.} \citep{finn2019online} introduced an online meta-learning approach based on the regret-based meta-learner. The approach performs MAML-style, which is a model agnostic meta-learning for adaptive deep learning, meta-training online during a task sequence. In this approach, there are meta-training and meta-testing phases in which we learn from input data in both phases training and testing phases. Even while testing, the meta-testing phase takes advantage of a training phase too. This ability make MAML looks powerful enough to yield a better result than other deep learning methods. Furthermore, MAML is one of the approaches that work best for few-shot learning (FSL) and one-shot learning (OSL) that enables classifiers to learn from a few samples of each category to predict unseen class labels with high accuracy \citep{mohammadi2020introduction}.
\\

\textbf{Further Challenges of Online Learning} 
Although online machine learning looks ideal for IoT healthcare applications, the challenging issue left in applying them with high accuracy \citep{mohammadi2020introduction}. One of the main challenging issues is catastrophic forgetting, where the new information prevents from learning what the model has already trained \citep{lee2020clinical}. This obstacle leads us to a significant failure in the classifier's performance while the new input data is being integrated or, regenerates a new model rather than keeping previous knowledge \citep{mccloskey1989catastrophic, mcclelland1995there}. Most of the applications for online learning in non-medical domains are less influenced by this limitation \citep{portugal2018use}. Online learning models in health care address a number of different challenges where it is needed multiple complex tasks.

A simple solution to address the catastrophic interference problem is to redo the training phase completely to regenerate the model every time new data are available, however, this process is computationally expensive and prevents from having real-time inferences \citep{lee2020clinical}.

\subsection{Federated learning}
Having heterogeneous IoT devices and associated different end-users ' (patients) information on them make it highly likely vulnerable for hacking while transferring information to an application server or when the data is restored on the server which raises data privacy and security issues. In this section, we aim to address these issues and investigate a solution that helps us solve them at an acceptable level. This solution must be able to train from isolated devices and integrate a model in a way that preserves users' information.

In a traditional machine learning-based approach, data gathered by IoT devices are uploaded to an application/data server and then trained models leveraged by machine learning algorithms. However, data owners (devices) have proliferated and data privacy gets important \citep{lim2020federated}, particularly in the medicine and healthcare field. In order to address the privacy requirement regarding individually identifiable, here, we propose to use federated learning, which was firstly proposed by Google \citep{konevcny2016federated}, which is a new approach to address this data dilemma, which is denoted as a challenging issue of training a high-quality shared global model with a centralized server with decentralized data distributed among a large number of devices or end-users \citep{xu2021federated}.

We propose to take advantage of federated learning (FL) which is a machine learning algorithm where the goal is to train a centralized rich model while training data remains de-centralized over a large number of users and devices where the network connections are unreliable and non-fast \citep{lim2020federated}. We use the federated learning algorithm for this situation where IoT devices independently perform an update to an input model (received from a central server) based on their local gathered data on demand and transfer this updated model to the central server, where the users-side updates are aggregated to generate/update a new global or generalized model. The common devices in this setting are mobile phones, in which data transferring and local sharing efficiency play an essential role in this setting.

\subsubsection{General Process of Federated Learning}
Federated learning is an approach to generate models, unlike other machine learning algorithms that need to have all data available as a central training dataset. In federated learning, a model is trained iteratively by aggregating collective models gathered from among multiple sources (devices: users information extracted from their cell phone). The users keep preserving their privacy and their local (training) dataset to themselves, but still may be able to participate in a shared federated learning process. Then, an aggregator would broadcast again the aggregated model to all the users and after that integrates again the updated model learning from the local training data in the devices \citep{lim2020federated}.
\begin{figure}[H]
    \centering
    \includegraphics[height=2.75in]{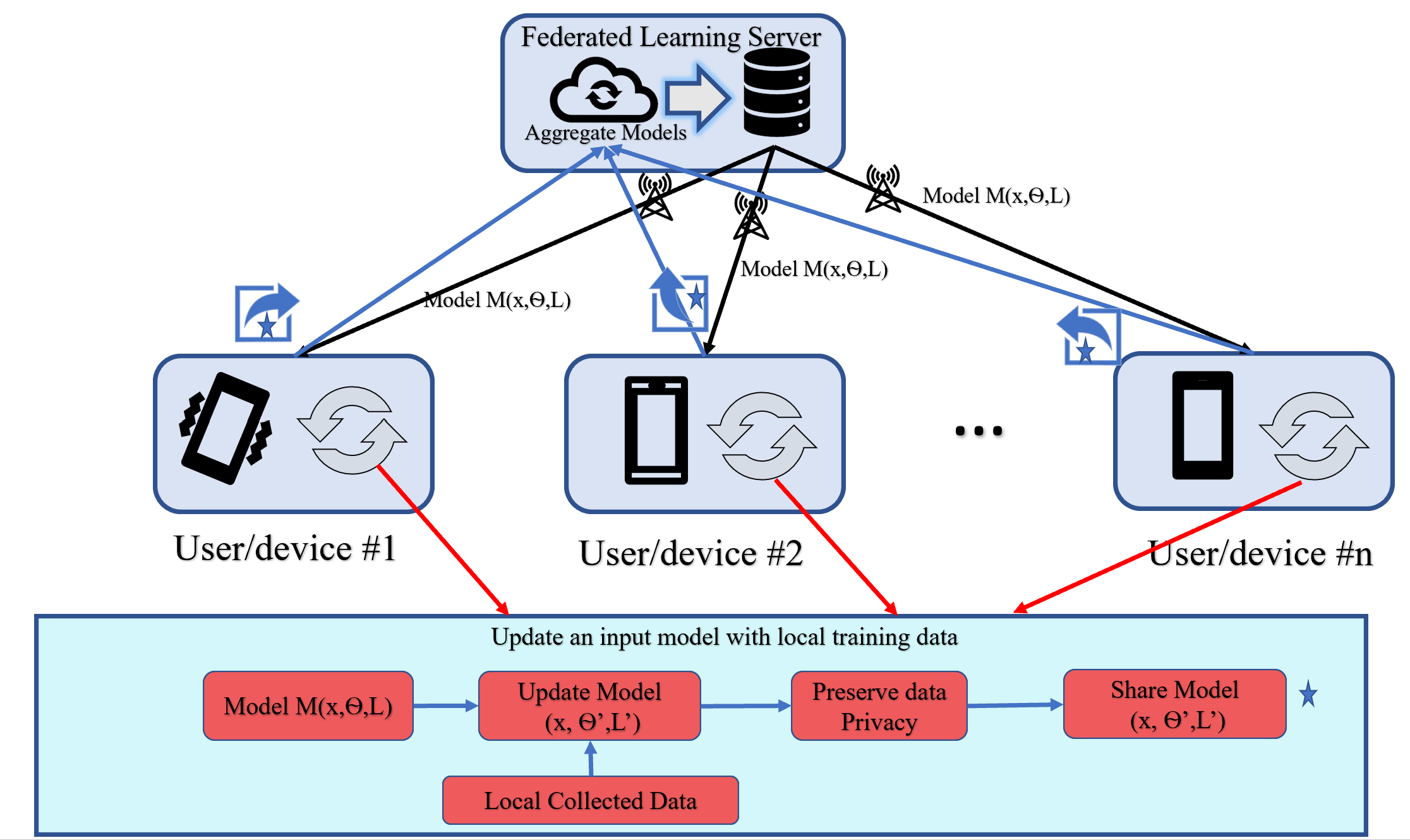}
      \caption{A general schema of Federated Learning }

    \label{fig:FL}
\end{figure}
We present a very abstract overview of the federated learning process in figure \ref{fig:FL}. This figure shows that a federated learning server communicates with all end-users / devices and gets them to check in the server and then start broadcasting the current model. The end-users / devices receive the model M with input data (x), parameters $\theta$ and loss function $\L$. Next, the device starts to update the model by learning from the local training dataset which is collected. The updated model M with updated values of parameters $\theta$' and loss function $\L$', per each device will send back the updated model. Finally, the server aggregates the received models from devices and then stores them in the server and keeps repeating this whole process. Having this process, enable specialists to monitor their patients accurately with updated information almost real-time. Furthermore, this FL prevents from sharing sensitive information. However, federated learning uses only cellphones as smart devices which is currently is one of the logistic limitations.

\subsubsection{Federated Learning Applications in IoT-healthcare }
Federated learning enhances the collaborative training of models such that the sharing raw data is none-sense. We need a federated learning system capable of preventing inference over both the messages exchanged during training and the final trained model while ensuring the resulting model also has acceptable predictive accuracy \citep{truex2019hybrid}.  
In \citep{wu2020personalized} researchers proposed an FL-based solution for learning a machine learning model namely, PerFit. PerFit works with a cloud-based architecture that provides computing power for IoT devices. In \citep{imteaj2020federated} the importance of the computation power is discussed in detail. Having the architecture well-structured provides a situation for IoT devices to unload their computing tasks due to efficiency and low latency requirements. As FL is communicating through different devices, servers, and the cloud, so models can be shared locally by preventing from
compromising sensitive data. The PerFit framework's learning process works mainly with three steps: 1) unloading tasks phase, 2) The learning phase, and 3) Personalizing phase. Researchers in \citep{wu2020personalized} evaluate PerFit's efficiency on a data-set called Mobile-Act, which centers on human activity recognition, which has ten different activities such as walking, jumping, and jogging.

In \citep{chen2020fedhealth}, a new FL-based framework, FedHealth,  using transfer learning was proposed for IoT healthcare applications. The transfer learning technique has been used in FedHealth to decrease the distribution divergence among a wide variety of fields. Furthermore, FedHealth uses a certain encryption algorithm to enhance the security of communicating current model updates between the aggregation server and end-users, and also vice versa. FedAvg has also been utilized as a federated optimization algorithm in the proposed FedHealth. There is a certain chance that the global model updated in FL centralized server is available to all end-users, but the personalized model in each user does not guarantee that performs well locally \citep{khan2021federated}. In order to handle this, to enhance the performance of the federated learning model for end-users, the researchers in \citep{chen2020fedhealth} leveraged transfer learning to train a personalized model for each user.

In  \citep{elayan2021deep}, researchers proposed a deep-learning based FL framework for decentralized healthcare applications that preserves data privacy and enhances data security in a decentralized architecture. They also proposed a scheme in which an automated training data acquiring process is applied. Additionally, they applied and evaluated the algorithm on skin diseases and leveraged transfer learning techniques to address the problem of lack of healthcare data existence in generating deep learning models.

\textbf{Further Challenges of Federated Learning} 
Although FL provides high data security levels and preserves data privacy, some posted attacks determined that simply sharing only local data during the training process, updated the model M does not guarantee sufficient data privacy \citep{truex2019hybrid}. Researchers in \citep{truex2019hybrid} proposed a hybrid approach namely, Privacy-Preserving Federated Learning, to yield a model with acceptable predictive accuracy and also is capable of preventing inference over both the messages communicating during the training process locally and the final trained model.

Furthermore, FL on IoT devices provides the following challenges: \citep{aledhari2020federated}: 

\textbf{Heterogeneity of devices:} IoT devices used for medical and healthcare issues and the ones used for general purposes are all different in terms of technologies and hardware such as the version of CPU, memory storage, and network connections bandwidth, storage capacity, and last but not least power. This could add more cost using FL, as numerous factors are required to get configured as fault tolerance. Furthermore, some devices may drop out of different learning processes due to various reasons like bad network connectivity and energy constraints.

\textbf{Statistical Heterogeneity and Heterogeneity of models :} In all machine learning, particularly FL, from collecting data point of view users' distributions of activity and settings play a fundamental role. Thus, different scenarios and settings due to very different physical features and behavior lead us to diverse data samples among devices in the FL that may cause an obstacle for making a rich model. Speaking of this, we end up having different diverse models in the server application to get them aggregate before broadcasting again.

\section{Conclusion}
The Internet of Things (IoT) is getting strong and powerful for the implementation of machine learning algorithms for the purpose of network monitoring and user activity management. However, traditional machine learning algorithms may fail while applying decentralized data collected from IoT devices. Since the nature of the algorithms is to get entire training data at once and generate a rich model which is capable of predicting unseen class labels during a test phase. Having distributed data prevent from yielding promising results using traditional machine learning algorithms. In this study, we discuss IoT, its pipeline, and its application in healthcare, and address the challenging issue ML algorithms may face. Finally, we investigate the process of data collection and its issues, and the impact of Big Data in IoT health, general IoT challenges, particularly the challenges of machine learning in IoT-healthcare, and associated novel solutions to address them.

\medskip
\bibliography{bib.bib}

\end{document}